# Very High-Resolution Bridge Deformation Monitoring Using UAV-based Photogrammetry

Mehdi Maboudi [1*], Jan Backhaus[2], Inka Mai[3,4], Yahya Ghassoun[1,5], Yogesh Khedar[2,6], Dirk Lowk[3,7], Björn Riedel[1], Ulf Bestmann[2] and Markus Gerke[1]

1 Institute of Geodesy and Photogrammetry, Technische Universität Braunschweig, Braunschweig, Germany; m.maboudi@tu-bs.de (M.M); b.riedel@tu-bs.de (B.R.); m.gerke@tu-bs.de (M.G.)
2 Institute of Flight Guidance, Technische Universität Braunschweig, Braunschweig, Germany; j.backhaus@tu-bs.de (J.B.); u.bestmann@tu-bs.de (U.B.)
3 Institute of Building Materials, Concrete Construction and Fire Safety, Technische Universität Braunschweig, Braunschweig, Germany
4 Institute of Civil Engineering, Chair of Robotic Fabrication of the Built Environment, Technical University of Berlin, Germany; mai@tu-berlin.de (I.M.)
5 Fraport AG Frankfurt Airport Services Worldwide, Frankfurt am Main, Germany; y.ghassoun@fraport.de
6 Autrik UG, Braunschweig, Germany; khedar@autrik.com
7 Department of Materials Engineering, Technical University of Munich, Munich, Germany; lowke@tum.de (D.L.)

* Corresponding author

## Abstract

Accurate and efficient structural health monitoring of infrastructure objects such as bridges is a vital task, as many existing constructions have already reached or are approaching their planned service life. In this contribution, we address the question of the suitability of UAV-based monitoring for SHM, in particular focusing on the geometric deformation under load. Such an advanced technology is becoming increasingly popular due to its ability to decrease the cost and risk of tedious traditional inspection methods. To this end we performed extensive tests employing an 18.5 m long research reinforced concrete bridge that can be exposed to a predefined load via ground anchors. Very high-resolution image blocks have been captured before, during and after the application of controlled loads. From those images, the motion of distinct points on the bridge has been monitored, and in addition, dense image point clouds were computed to evaluate the performance of surface-based data acquisition. Moreover, a geodetic control network in stable regions is used as control information for bundle adjustment. We applied different sensing technologies in order to be able to judge the image-based deformation results: displacement transducers, tachometry and laser profiling. As a platform for the photogrammetric measurements a multi-rotor UAV DJI Matrice 600 Pro was employed, equipped with two RTK-GNSS receivers. The mounted camera was a PhaseOne iXM-100 (100 MP) with an 80 mm lens. With a flying height of 30 m above the terrain this resulted in a GSD of 1.3 mm while a forward and sideward overlap of 80% was maintained. The comparison with reference data (displacement transducers) reveals a difference of less than 1 mm. We show that employing the introduced UAV-based monitoring approach, a full area-wide quantification of deformation is possible in contrast to classical point or profile measurements.

**Acknowledgments:** The Project "Optical 3D-Bridge-Inspect" is funded by the Deutsche Forschungsgemeinschaft (DFG, German Research Foundation) – Project Number 501682769 as part of DFG's Priority Programme 2388 "Hundred plus". The authors would also like to acknowledge support by the Open Access Publication Funds of the Technische Universität Braunschweig.

# 1 Introduction

Infrastructure inspection, particularly of critical structures such as bridges, is of paramount importance in ensuring public safety and the longevity of transportation networks. For the majority of bridges worldwide, standard regular inspections are envisaged, e.g. [1], [2], [3], [4], [5]. In the United States, The National Bridge Inspection Standards (NBIS) mandates a routine inspection of the bridges, every two years [6]. In Germany, the manual, cyclical inspection of civil engineering structures according to DIN 1076 2009 [7] has been state of the art since 1930. Currently, damage detection and classification are based on visual field examinations and on the results of basic non-destructive and/or semi-destructive tests using various types of techniques [8], [9], [10], [11], [12]. The results of these inspections form the basis for decisions and actions derived from the results of bridge inspection within the framework of maintenance management [13]. In some cases, these inspections are already accompanied or supplemented by a sensor-based condition assessment, the so-called Structural Health Monitoring (SHM) [14], [15], [16], which is an efficient solution especially when implemented in sensor networks [17]. The aim is to evaluate the structural health of a structure or performance on this basis [18], which is not plausible by means of inspection-based monitoring only.

In SHM, behaviour and response of the infrastructure to the anticipated loading should be quantified and monitored. Especially, displacement has been directly used as a safety index in structural design codes all over the world, e.g. [19], [20], [21]. In normal operation conditions, short- and long-term device-based monitoring are distinguished [14]. The short-term monitoring usually lasts a few days, and temporarily installed sensors are used for data acquisition, for example, for predefined proof loads before the opening of a structure to the public or after completing a major rehabilitation work. The long-term monitoring requires a permanently installed mounted system in order to gain information for example about the structure’s reaction to impacts such as traffic loads or environmental impacts due to temperature, humidity or wind. In this paper we use the term displacement when referring to individual observations which only provide information about a single position on the structure. In contrast, deformation refers to the overall shape modification of the object.

During the last decades, the technology and methods for close-range, contactless, optical sensing developed considerably. Contactless methods are needed if the direct accessibility of the structure is not possible or desirable. This concerns not only terrestrial laser scanning, where small volumes can be sampled at very high-resolution and accuracy, but also unmanned aerial vehicles (UAVs/drones) which enable capturing high-quality image blocks in flexible layouts [22], [23]. Using state-of-the-art photogrammetric technology, including automatic sensor orientation (structure-from-motion) and dense image matching, surface geometry can in principle be reconstructed in sub-pixel resolution and accuracy [24]. While the mentioned conventional in-situ SHM measures deformation on pre-defined points at the structure, photogrammetric methods in combination with high-resolution UAV-based image blocks would enable an area-wide analysis of deformation.

To this end, we examined the research question of whether high-resolution UAV-based photogrammetry is able to generate results that are accurate enough for deformation inspection in SHM by comparing its results to those measured with multiple different classical SHM and surveying techniques. Furthermore, we analysed if such a holistic approach generates additional value for the inspection regarding completeness and area-wide analysis. It should be accentuated that while some studies e.g. [18], [25] measure the bridge deformation under dynamic load, others address the problem under static load [26], [27], [28], [29]. The advantage of static load experiments is that especially area-based measurements can be done more accurately, since for common techniques like laser scanning or multi-image photogrammetry, a static scene is a pre-requisite. In our case study, we therefore deformed the research bridge “*Concerto*” [30] with a defined static load in well-defined loading and unloading steps. We measured the corresponding deformation using four contact and contactless methods, which differ in temporal and spatial resolution, accuracy and setup effort:

- Conventional displacement transducers
- Tacheometry
- Laser scanner profiling, and
- UAV-based photogrammetry

It should be accentuated that in this paper "deformation" refers to overall change of the shape of the whole object, while displacement refers to change in single point's coordinates. Therefore, we use the term displacement whenever a single point is measured e.g. using transducers. Moreover, the term UAV might refer to unmanned, unattended, or uncrewed aerial vehicles. Also, there are different terms for UAV such as drone, UAS (Unmanned Aerial System), RPAS (Remotely Piloted Aircraft System) that are being used in the literature, interchangeably [24].

This paper is organized as follows: In Section 2, the state-of-the-art methods for measuring deformations of large urban structures with the main focus on the bridges are discussed. Section 3 describes the materials and methods. Here we provide detailed information about the test bridge "*Concerto"* in our case study and explain the techniques employed for deformation measurement. In Section 4, experimental results of deformations of the loading and unloading phase are presented and discussed. Section 5 concludes the paper and highlights our recommendations for future works.

## 2 Related Works

In the presented research, the structural performance of a bridge due to loading and unloading is of interest. There are multiple ways to measure deformation of a bridge and we divide them into 3 groups: point-based, profile-based, and area-based deformation measuring methods.

### 2.1 Point-based Methods Using Displacement Transducers and Total Stations

Generally, point-based methods are relatively accurate, as for example inaccuracies of displacement transducers are usually better than 1% of the actual displacement. However, as they only determine the displacement of some distinct points, many independent measurements per epoch are needed to not only extract the magnitude but also the shape of the deformation e.g. along the bending line of the bridge. Point-based dynamic monitoring data has been used extensively in the past, e.g. to derive a reliability index of a structure and herewith enable a buckling check of a bridge [18], or to enable the proof of fatigue resistance of prestressing steel at coupling joints of bridges [31].

A point-based method that is often applied in bridge monitoring, is using displacement transducers. We consider it the most accurate measurement in this case study, but as each of these sensors can only be mounted on a distinct point where a ground bearing is present, the results are spatially very sparse, while temporally very dense.

Another well-established method for monitoring the behaviour of infrastructure objects for many decades is the use of high precision ground surveying. With the development of robotic and automatic tracking functionality of the total stations, one can continuously observe a predefined number of discrete and expressive points. However, due to the point-by-point discretization, only a rough approximation is obtained regarding the creation of a bending line.

In the framework of an SHM project, the discrete points need to be split into two separate groups of points according to the approach of stable reference points and object points, which are subject to displacement [32].

### 2.2 Profile-Based Methods Using Terrestrial Laser scanner (TLS)

The use of TLS has shown notable success in monitoring large civil structures such as bridges, buildings and tunnels [28], [33], [34]. By using TLS, one can scan a profile along the bridge before and after a controlled loading of the bridge thereby extracting a bending line of the deformation. However, due to the geometrical setup, the density and accuracy of the resulting point cloud reduce with distance from the scanner. In [35], authors employed TLS-based data acquisition to monitor the bending behaviour of a bridge using a laboratory setup. The authors could show that TLS-based measurements are of sufficient quality to describe a relation between bending behaviour and failure of the structure in a loading stage. In the case of profile scanning, the derived deformation line is only indirectly related to the surface measurements. An advantage of the chosen setup is that TLS profiles can be measured at high frequency, leading to higher precision of the finally acquired bending lines when those measurements are averaged. Three approaches, namely point-to-surface (P2S), point-to-cell (P2C) and cell-to-cell (C2C), are introduced in [36] to measure the deformation of a bridge in Ireland using laser scanning data. Investigating the pros and cons of each of the approaches, the mean estimated values of the deformation employing different methods are reported to be not more than 3.2 mm. A phase-shift TLS is employed in [27] to measure the static deformation of a 33-year-old concrete bridge in Switzerland. The load test was performed in four epochs and the TLS-based deformation result was compared to a precise levelling. Although differences up to 3.5 mm are reported, the mean values of the vertical displacements are similar (less than 1 mm difference). The authors suggested considering TLS-based deformation monitoring as a complementary survey method for load tests, together with precise levelling. Lõhmus et. al. [37] utilized a time-of-flight (ToF) TLS to measure the vertical deformation of two bridges. They tested one highway bridge and one railway overpass using TLS data, and verified the results with dial gauges, tacheometry and precise levelling. The results confirm the suitability of TLS data for deformation monitoring in a range of few millimetres.

Although TLS is fast, non-contact, and delivers dense 3D data even from most inaccessible parts of structures, the acquired data could be strongly affected by material properties, measurement planning (coverage, incidence angle, etc.), and spatial and temporal co-registration of the captured data [29], [34], [38], [39]. Furthermore, TLS measures locally, and without proper connection to external reference and in a good network setup, the point cloud might be deformed. Therefore, most research on this topic conclude and suggest TLS-based deformation monitoring as a complementary survey method for load tests, together with high-precision geodetic survey [27], [28], [29], [37]. Further details on various aspects of using TLS in deformation monitoring can be found in literature [33], [34], [40].

## 2.3 Area-Based Methods Using UAV-based Photogrammetry

In area-based methods, various vision-based systems could be considered. For example, [41] reported the effectiveness of a combined sensing system which consists of a consumer-grade camera (mounted on a tripod for video recording) and accelerometers for bridge deformation monitoring. The UAVs are inexpensive, accessible, fast, and flexible data acquisition platforms and have already proven themselves as powerful tool for accurate mapping and 3D reconstruction purposes [39], [42], [43]. Therefore, they are considered as the main image-capturing platforms in this paper. Various aspects (such as the number and distribution of ground control points (GCPs), UAV flight schema, and utilized sensors, just to name a few) can affect the quality of 3D reconstructed models. The interested reader can refer to [44], [45], [46], [47], [48] for more information about the parameters that affect the quality of UAV-based 3D reconstructed models. UAV photogrammetry is already employed for the inspection of civil structures. Most notably, structural damage identification is one of the main trends in this field [49], [50], [51], [52], [53], [54]. Though 3D photogrammetric data (point cloud) are also used for bridge inspection [55], damage identification approaches are mainly based on 2D information (image processing, mainly in one epoch). Nevertheless, deformation analysis of large structures like bridges demands 3D data processing with the main focus on vertical deformation in various epochs.

Deformation of road sections using UAV-based photogrammetry is reported in [39]. Different flight configurations were tested, and the authors confirmed that with an optimal GCP configuration, point accuracies of 1.5 mm (40 m flight altitude) to 3 mm (60 m flight altitude) are achievable. Derived area-wide surface deformation is evaluated using profiles from levelling and showed discrepancies of about 5 mm RMSE (Root mean Square Error).

In order to quantify the deformation of short segments (one meter) of a steel girder bridge, in [56] UAV data are not georeferenced and processed locally (after co-registration of the point cloud from different epochs). The effect of various co-registration algorithms is reported, and the results are compared to string potentiometer readings. Although the discrepancy of around 2 mm between UAV and ground truth data is reported, the scalability of the outcomes to larger bridge segments should be further investigated. The influence of various parameters on the quality of 3D reconstruction using UAV images for bridge inspection is investigated in [42]. The effects of partial image overlap, different flight altitudes, local camera-object distance, and GCPs and checkpoints arrangements are studied. Employing a small concrete beam (1.0 m × 0.1 m × 0.2 m), the authors imposed an artificial deformation on the object and compared the proposed optimal UAV data capture and 3D reconstruction results with tacheometry measurements. Although the results are promising, no area-based dense matching was employed and the size of the object is rather small, so conclusions about obtainable accuracy in real-world scenarios need further investigation. There are also some other research works that employed and compared different sensors for bridge deformation monitoring. For example, in [57], three different sensors (satellite radar interferometry (InSAR), UAV-based LiDAR, and vehicle-mounted mobile laser scanning) were utilized for the deformation monitoring of a bridge. The authors reported the pros and cons of those approaches for network-wide and also single-bridge inspection.

Since point cloud based deformation detection relies on the accuracy of point clouds comparisons, one important step is the co-registration of data from different epochs [58]. One approach is to compute the relative deformation (difference between, e.g. UAV-point clouds of different epochs) and compare it to the ground truth deformation value (for example, string-potentiometer in [56]). Here, the point clouds co-registration problem arises and could be tackled as reported in [56]. Employing conventional approaches like C2C co-registration that minimizes the difference between two point clouds, might affect the realization of real deformation. Another approach is direct georeferencing (onboard multi-sensor system) or indirect georeferencing (using GCPs) of the data and avoiding direct co-registration of different epochs. Here, the georeferencing accuracy would be important and can affect the deformation analysis reliability. In our research, we combine both direct and indirect georeferencing to avoid pitfalls of the C2C co-registration and also to benefit from the advantages of both georeferencing methods. For more details on co-registration of point clouds and photogrammetric data, the interested reader can refer to [59], [60].

# 3 Materials and Methods

## 3.1 Experimental Bridge “Concerto”

For long-term testing of innovative SHM methods, the experimental 2-span plate girder bridge "*Concerto*" (Figure 1) was built in 2005. The bridge has a total length of 18.5 m, a horizontal plate width of 4.0 m and a vertical web height of 0.8 m [30], [61], [62]. Figure 1 shows in the upper part the location of the supports (A, B), the ground anchors (C), the transducers (C, D) and the cantilever arm (E). In the lower part of Figure 1 the instrumentation and reference point signalization are depicted.

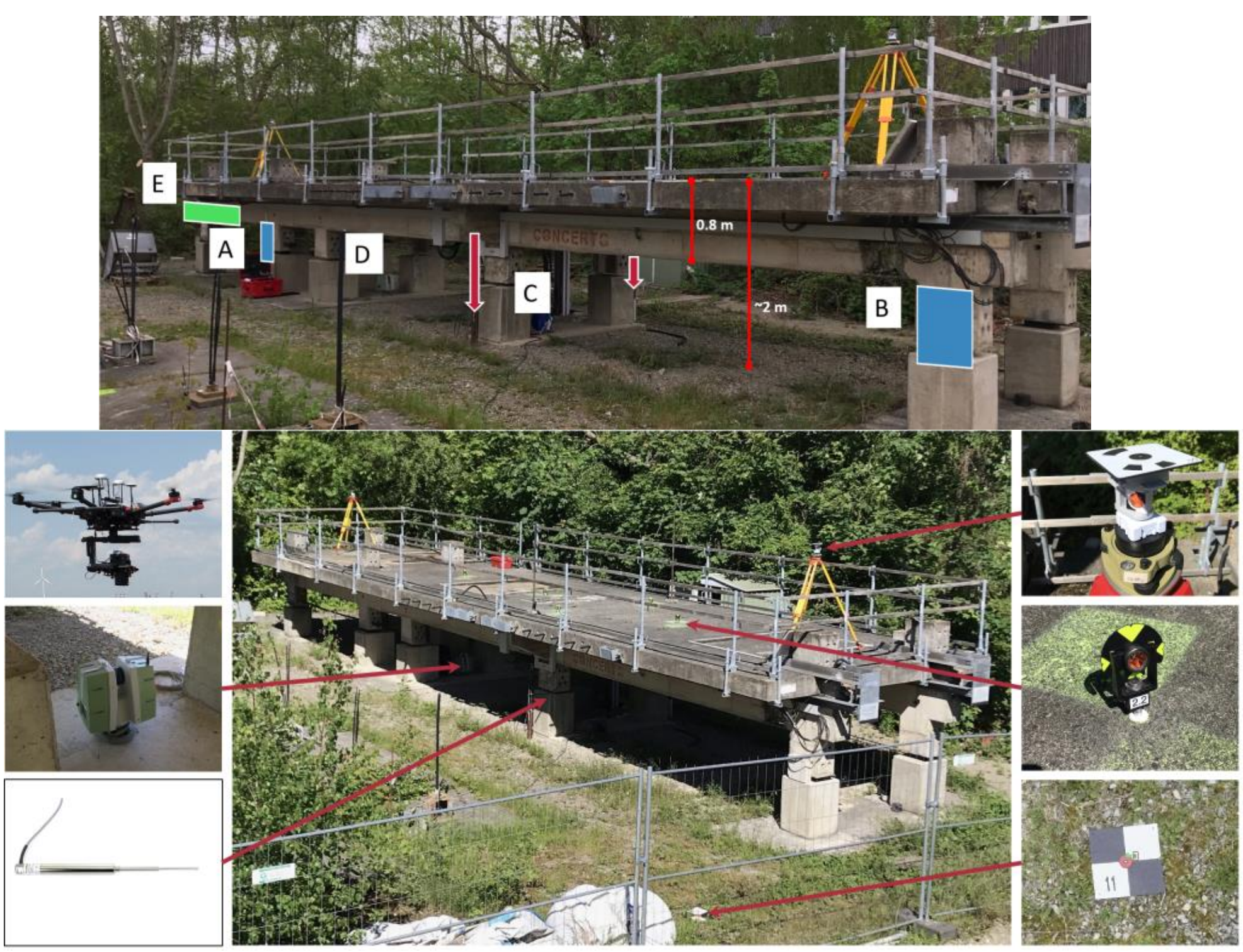

Figure 1. Experimental Bridge “*Concerto*”. Upper part: indication of support (A, B), ground anchor (C), Deformation transducer positions (C and D) and cantilever arm (E). Lower part: instruments (UAV, laser scanner and transducer) and signalization of reference points

In Figure 2 a section of the “*Concerto*” bridge is shown, overlaid with assumed deformation curves. The structure is supported at two points along the bridge (position A and B) and is cantilevered at one side (area E). In addition, it is possible to put additional local loading on the bridge via ground anchors (position C). The blue line shows the shape of the bridge in an unloaded state. When the bridge is loaded at position C, we assume, that it will lower there, while the cantilever arm E is expected to rise. The green line shows the assumed deformation qualitatively. When loading the experimental bridge in three steps, a load is applied, and simultaneously, the deformation is measured with displacement transducers continuously.

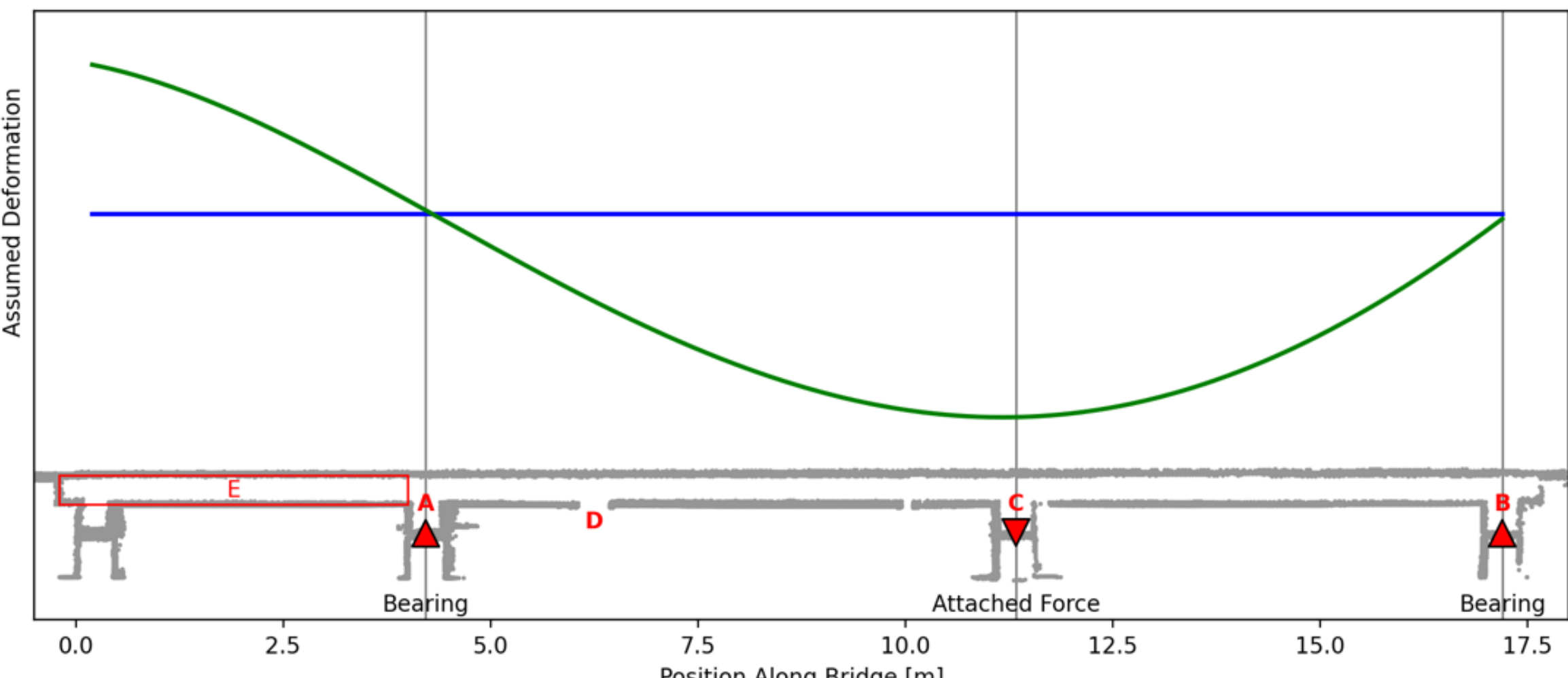

Figure 2. Schematic Shape of “*Concerto*” with bearings (A & B), attached force (C) and cantilever arm (E). Green line shows the assumed deformation under load.

In the presented experiments, the loading of the structure "*Concerto*" is carried out by two Dywidag single-bar anchors (diameter 26.5 mm, ST 950/1050), which are anchored into the ground at position C. The ground anchors are unloaded and loaded by means of a heavy-duty lifting cylinder. The loading is conducted in three steps up to approximately 100 kN. Load cells are installed at the load setting device to control the load applied to the bridge. The deformation behaviour of the bridge is monitored using displacement transducers at four positions (two approximately in the middle of the field where the load is applied (position C) and two that are closer to support A (position D). The positions of displacement transducers are marked as yellow triangles in Figure 4.

## 3.2 Deformation Experiments and Epochs

We conducted several deformation experiments at the experimental bridge “*Concerto*”. Because it had been loaded for 17 years prior to our experiments (defined as epoch 1), we first had to unload it (epoch 2), see Figure 3. After waiting for one day, we started with the unloaded state (epoch 3) and reapplied a load, resulting in epoch 4. After waiting for 20 days, we again measured the deformation (epoch 5). Table 1 shows a list of these epochs with a description of the according loading state. In this paper we focus primarily on the process of reloading the bridge between epochs 3 and 4 and after the unchanged load in epoch 5. Epochs 3 and 4 resemble a controlled loading state and the time between these epochs is less than 2 hours. Epoch 5 is used to analyse the behaviour after constant load conditions. The initial unloading started with an unknown loading state from years ago, and thus epochs 1 and 2 are not discussed in the following. Figure 3 shows the timeline and sketches of the assumed deformation characteristics of the bridge in each epoch.

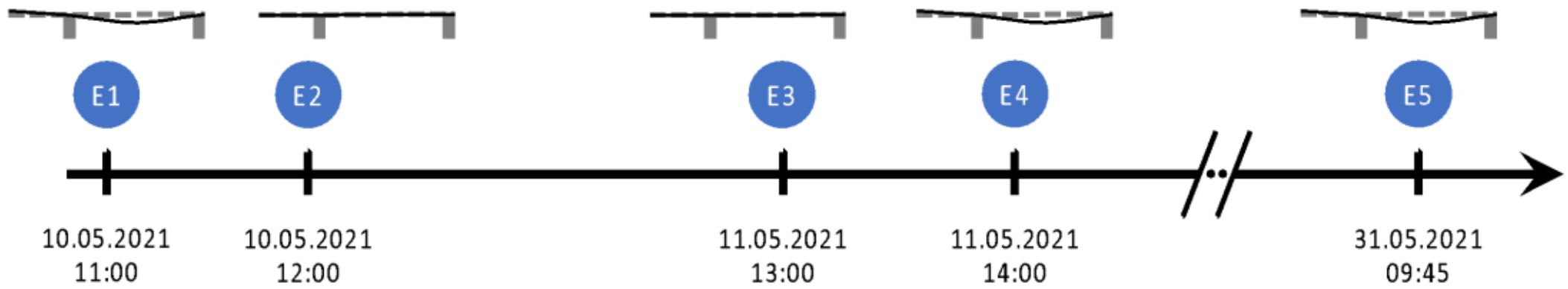

Figure 3. Timeline of the epochs in our experiments. Top sketches show the assumed deformation of the bridge in the corresponding epochs qualitatively.

Table 1. Overview of the epochs of our experiment.

| Name | State | Description |
|---|---|---|
| Epoch 1 | loaded | Morning of the first day. Initial state. Load was applied 17 years ago. |
| Epoch 2 | unloaded | Noon of the first day. State after unloading the bridge. |
| Epoch 3 | unloaded | Noon of the second day. Bridge still unloaded and relaxed overnight. |
| Epoch 4 | loaded | Early afternoon of the second day. Bridge under load again. |
| Epoch 5 | loaded | Morning 20 days after reloading. Bridge still loaded. Long-term comparison. |
| Epoch 1□ Epoch 2 | unloading | Process of initial unloading. Not presented in detail in this paper. |
| Epoch 3□ Epoch 4 | loading | Process of loading from unloaded to controlled loading state. |
| Epoch 4□ Epoch 5 | constant load | Stable loading condition |

In the experimental area, we created a local surveying reference network, containing points on the upper side of the bridge and next to it on the ground. The bridge points were measured with a total station in every epoch, cf. 3.3.1. At the same time, we also created a profile using a TLS (cf. 3.3.2). In each epoch, we also employed an automated mission with a UAV system, taking aerial images of the bridge. This process is described in detail in subsection 3.3.3. It is also worth mentioning that we spent 2 hours on selecting and marking the reference points, total station observations supported by GNSS measurements took 4 hours, UAV flight took 4 minutes.

## 3.3 Methods of Deformation Measurement and Implementation

### 3.3.1 Transducers and Tacheometry

When loading the experimental bridge in three steps, a load is applied and simultaneously the local displacement is measured with displacement transducers continuously (HBM, W20TK, displacement measurement inaccuracy of less than 1% of the actual displacement according to DIN EN ISO 9513, and measurement frequency = 0.02 Hz). These transducers are mounted in the middle of the field, i.e. where the ground anchors are located (position C Figure 2) and at the edge of the field (position D in the same figure). As at each cross section of the bridge (position C and D) two displacement transducers are installed (Yellow triangles in Figure 4), the reported values are the interpolated projection of each pair of two transducers on the bending line (centreline of the bridge). Because of the direct method of measurement, we consider these values as ground truth for our experiment.

In this investigation we built up a hierarchical surveying network with two groups of points (as mentioned in section 2.1). As depicted in Figure 4, the stable reference point field outside the bridge structure consists of two instrument stations (point number 10, 40), as well as six GCPs (1, 2, 3, 5, 7, 8) with target signs for the UAV photogrammetry and two points (R1, R2) at a building close to the bridge for the linking and stabilization of the reference network. All ground points were marked out in the ground using tubes with centring caps and the target on the building were equipped with reflective markers. The seven object points (20, 21, 22, 22, 23, 24, 25 and 30) on the bridge, which further serve as checkpoints for the photogrammetry, were marked on the bridge surface with screw-in dowels and were equipped for the measurements with so-called mini prisms. The position of the object points is based on the longitudinal axis of the structure, as well as on the positions of the force application to the structure (points 22 and 23). The two outer bridge points (20 and 30) were additionally used for the initial measurement (epoch 1) as instrument locations. This initial measurement, so called zero-measurement was carried out some days before any experiment started. The complete network design is shown in Figure 4.

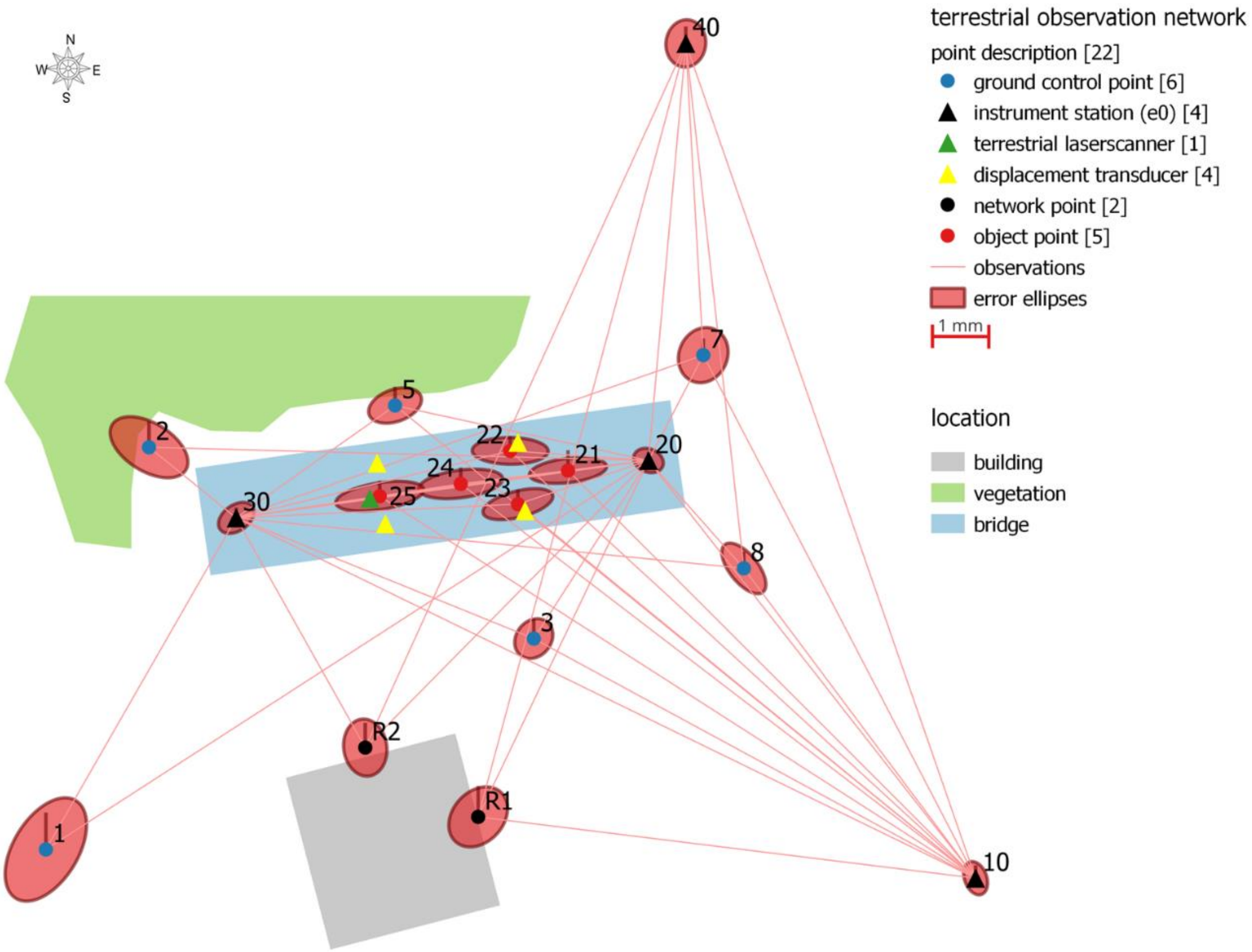

Figure 4. Sketch of the network design and overview of the sensor locations

The objective of the initial network measurement is to create a stable, geometric reference field around the object of interest that meets the requirements of high accuracy and reliability. For all reference and object points, real-time kinematic (RTK) GNSS observations were carried out to ensure a localization and datum definition of the point coordinates in the UTM coordinate system with ellipsoidal heights for further aerial surveys. As the tacheometric measurements have a much higher accuracy, the GNSS-RTK points were only included in the network survey with low weights to avoid negative influence on the inner network geometry and to deliver a global reference. The further geodetical observations were carried out with a high precision total station using points 10, 20, 30 and 40 as instrument stations. The six GCPs were only equipped with simple plumb rods with prisms for this measurement. Two to three full sets of observations per station were measured. The network adjustment results of this initial campaign led to a standard deviation of all points in North and East direction of better than 0.9 mm and for the height component better than 0.7 mm.

### 3.3.2 Terrestrial laser scanning

Finding an optimal position of the device was driven empirically to capture the location of the largest deformation, close to the axis of ground anchors, at a good angle of incidence. It was also necessary to consider that the distance between the device and the underside of the bridge was very short (<1.50 m). This short distance causes a strong measurement noise which had to be reduced at the location of the largest deformation. That is the reason why the scanner was placed approximately 5 m away from the axis of the ground anchors, close to the transducer at position D in Figure 1.

The pillar plate was placed in the ground (in the middle of the width of the bridge), and the used tribrach stayed in forced centring the whole time of the experiments so that the risk of a shift in the scanner's vertical axis should be minimal. This means it was possible to avoid errors from localization procedures since all measurements are, per definition, made from the same position. The chosen setup can be seen in Figure 5.

The scanner's azimuth orientation needed to be renewed before every epoch. With a fixed point of orientation at the end of the bridge (West), which was also centred in the transverse axis of the bridge, the orientation could be computed before every measurement. Finally, an orientation uncertainty of 3 cm between the single derived profile lines was obtained. However, since we have a continuous measurement, the computed average was regarded as good enough for next steps. The profile lines at every epoch were captured with a resolution of 3 mm @ 10 m and a frequency of 50 Hz.

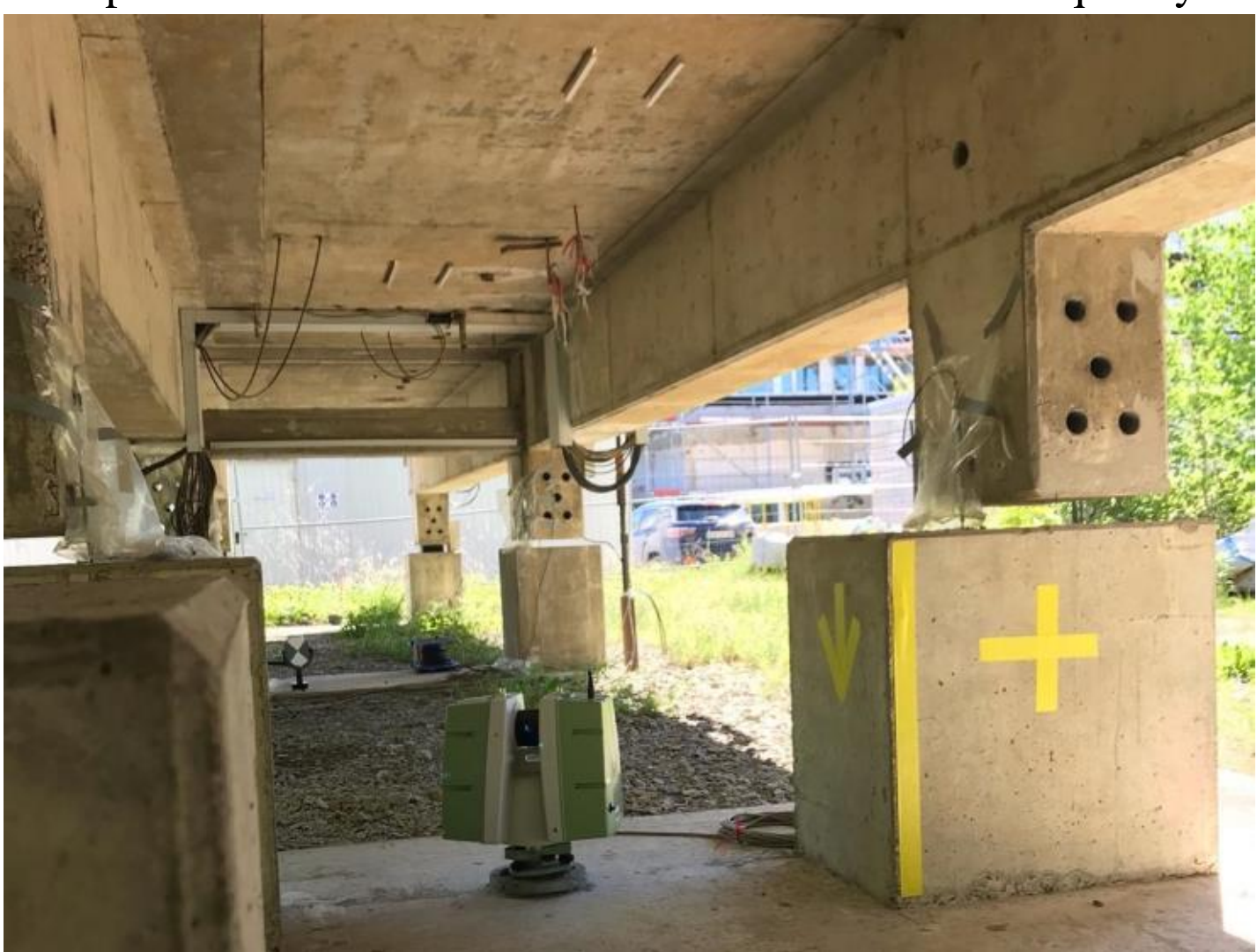

Figure 5. The TLS (Leica P20 Laser scanner) mounted under the bridge, close to position D in Figure 1

It should be highlighted that our experiment was performed on a test bridge which its height (less than 1.5 m : see Figure 1) is by far less than the functional bridges. This brings some limitations on TLS setup that leads to non-optimal results. The authors believe that in real cases where there is enough space underneath the bridges, better results could be achieved using terrestrial laser scanning.

#### 3.3.3 UAV-based photogrammetry

Our approach to area-based deformation measurement using UAV-based photogrammetry is based on capturing high-resolution images from a nadir-mounted camera on a UAV with high overlapping rates. The UAV shall fly on an automated zig-zag mission, ensuring constant overlap and altitude.

Structure from Motion (SfM) is the process of reconstructing 3D structure from its projections into a series of unordered images, given that the images are taken from various viewpoints with proper overlap [63], [64]. Utilizing state-of-the-art SfM implementation of *Agisoft Metashape*, where ground control points are properly included into the bundle adjustment and multi-view-stereo methods, we generate a dense point cloud of the bridge’s top surface and derive a digital elevation model (DEM). The implementation works according to the approach by Furukawa and Ponce [65], which is a step-wise densification of the sparse point cloud stemming from the bundle adjustment (tie points). Here the multi-views are exploited to reach higher reliability and accuracy compared to stereo matches. In the commercial implementation, however, some refinements of the original approach got implemented which are not fully disclosed. By using the difference between the DEMs of two epochs with different controlled loading states, we aim to extract the bending lines of the bridge along multiple axes. We have used such a high-resolution reconstruction pipeline before monitoring rail positions in an industrial context, where we reached a vertical precision of below 2 mm [66].

For the experiments described in this investigation, we used a commercial DJI Matrice 600 Pro multicopter carrying a PhaseOne iXM-100 camera, which is shown in Figure 6. This UAV is equipped with a differential RTK system that allows positioning of the drone and referencing of the images with centimetre-level accuracy.

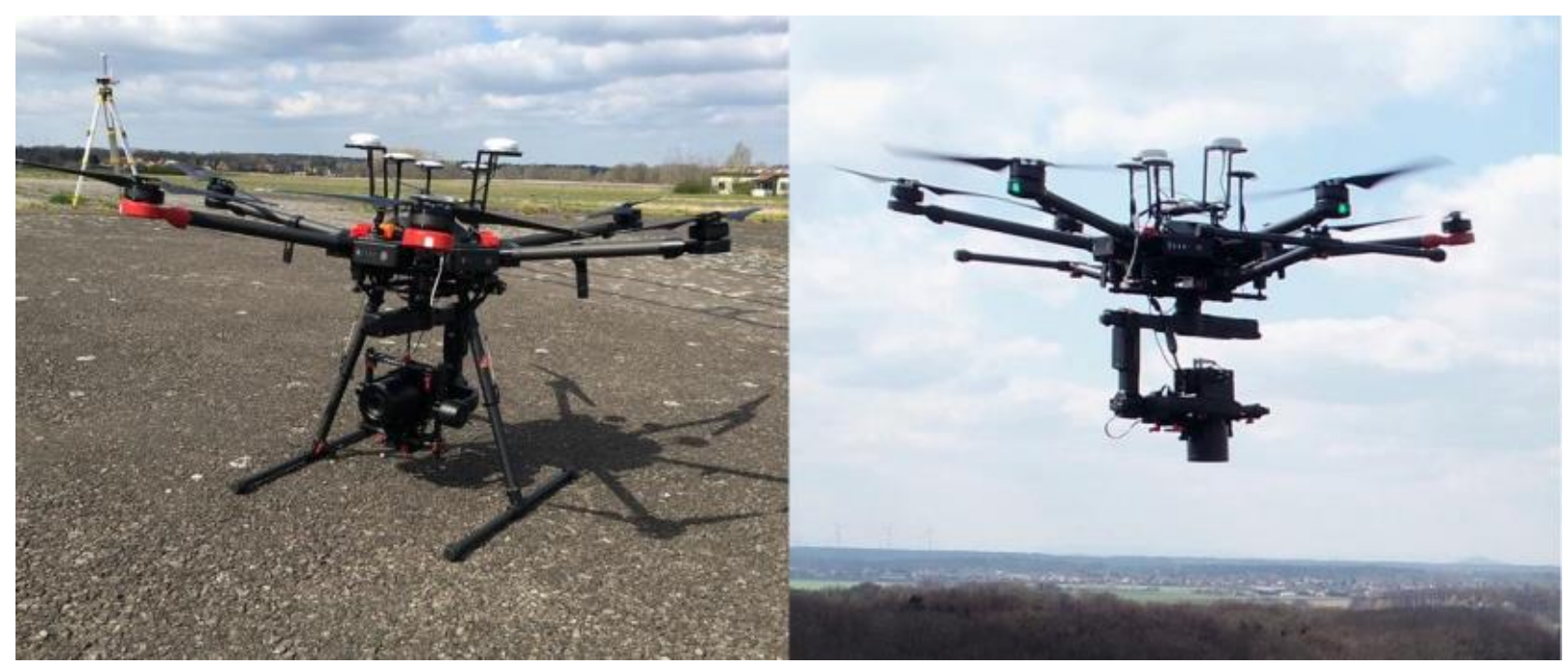

Figure 6. DJI M600 Pro with PhaseOne iXM-100 UAV mapping system on the ground (left) and in-flight (right).

The iXM-100 camera with 100 megapixels and pixel pitch of 3.45 µm, and an attached 80 mm-lens is stabilized during flight with a 3-axes gimbal. From an altitude of 30m above ground level, this setup generates images with a GSD of 1.3 mm. Flight planning took place with a self-implemented *Inspekt GS* app for Android devices, which allows a fine-grained input of flight parameters. For this case study, we took nadir images with an overlap of 80%, resulting in 55 images per epoch. A lower overlap would result in fewer images. It would reduce the computation time significantly, but from our experience with the setup in comparable applications [67], 80% is a good compromise between computation time and reconstruction accuracy. Nevertheless, we tried a higher overlap of 90% in some preliminary tests for this experiment and did not notice any better results but, instead, a drastic increase in computation time, as the number of images was four times more than 80% setup. The flight planning process in the app can be seen in Figure 7.

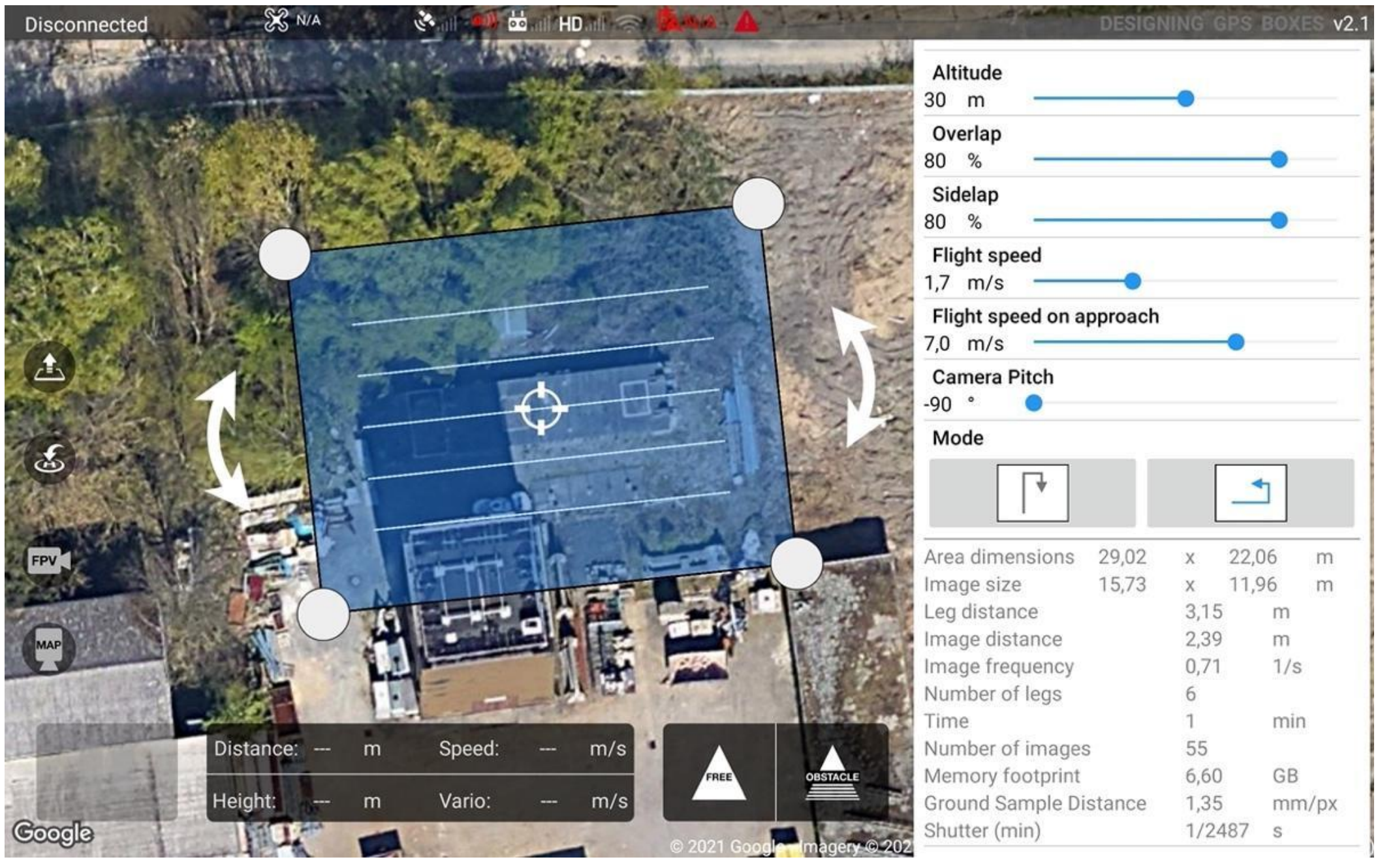

Figure 7. Flight Planning in Inspekt GS Android App.

In every epoch, one flight was carried out, images were taken and dense reconstruction of the bridge and its surrounding environment was calculated using the commercial software *Agisoft Metashape* with the highest setting of the parameters. To scale and georeference the dense cloud, 6 GCPs were placed around the bridge and assumed to be static between all epochs. See section 3.3.1 for details on the network measurements. The points were marked using non-coded cross-shape targets and manually selected in all the images. The points on the bridge itself, which moved with it, were used as check-points (CPs) in each epoch. This means, they were not included as observations in the bundle adjustment. In order to be able to measure those points with both, the total station, and in the images, the signalization had to be changed: for the total station measurements, reflectors were added (see Figure 1, lower part), which were then removed for the image-based observation, i.e. only the bolt was visible in the images.

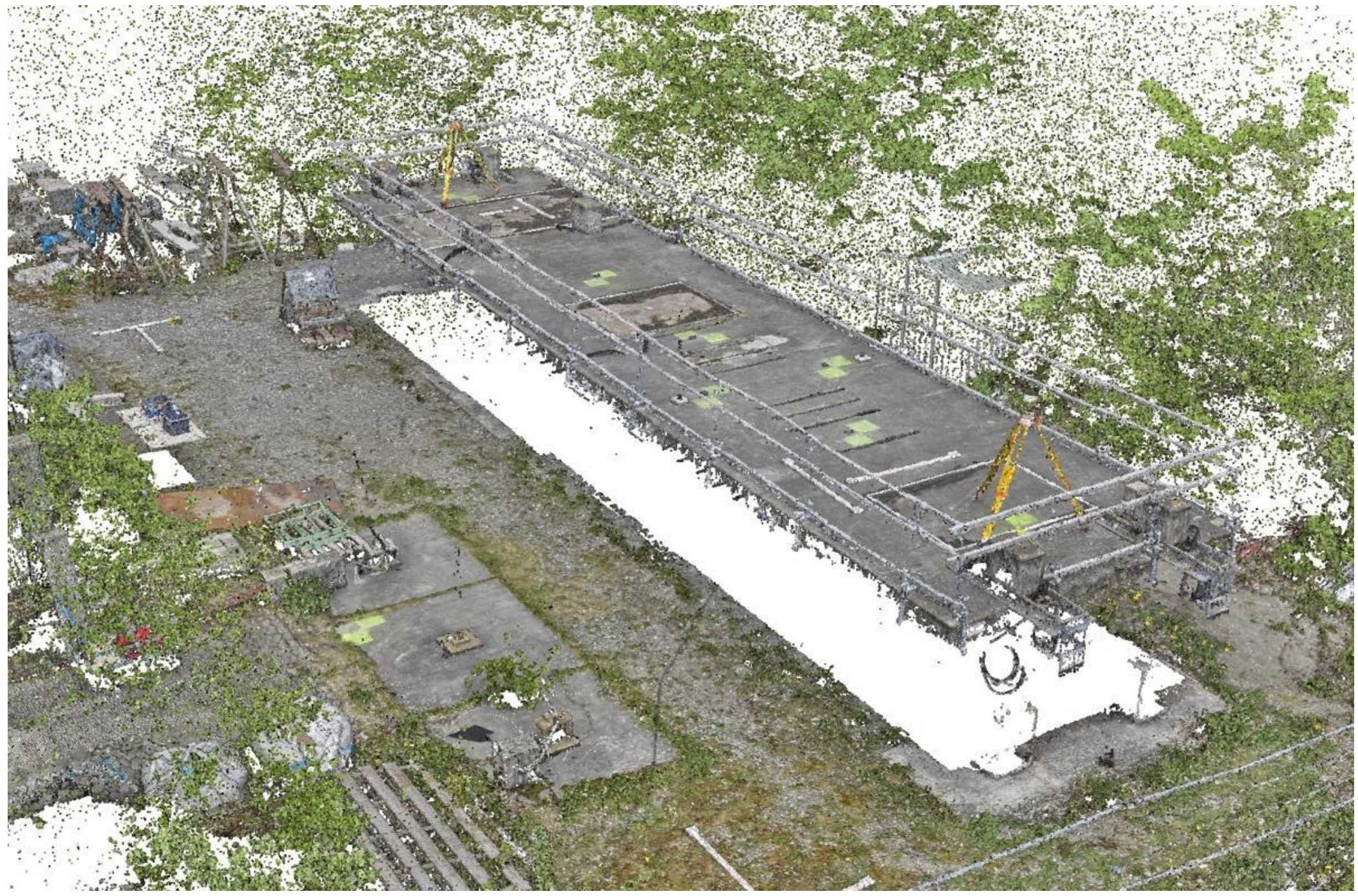

Figure 8. Dense reconstruction of the "*Concerto*" bridge at epoch 4.

Based on the dense clouds of each epoch, digital elevation models of the surface were extracted with the point spacing of 1.3 mm (almost 1 point/pixel) and used for further processing. Figure 8 shows the dense cloud of epoch 4 as an example. Only the top surface of the bridge is visible because of the nadir setup, but it is the relevant surface to determine the vertical deformation. Each of the dense clouds consists of around 6 million points.

# 4 Experimental Results and Discussion

In this investigation different methods for measuring the deformation of the "*Concerto*" experimental bridge under defined loads are examined. Section 4.1 describes the detailed results of each method, whereas section 4.2 compares the results of those methods.

## 4.1 Results of the Examined Methods

The results of the displacement transducers are considered ground truth in our experiments, as this method works directly on the bridge and next to the point where the force is applied. Section 4.1.1 shows the results of these transducers for the deformation during the period of load application (epochs 3 to 4) and also covers the long-term deformation behaviour of the bridge (epochs 4 to 5). In Section 4.1.2, the displacements of the checkpoints on the bridge during the loading process are shown, as they were measured with the total station. The profiles of the TLS are presented in section 4.1.3, which show the bending line of the bridge. Section 4.1.4 describes the results that can be achieved using UAV-based photogrammetry.

### 4.1.1 Results of displacement transducers

In Figure 9 the recorded loading and corresponding deformation behaviour of the experimental bridge are shown. In the first loading step, the loading cells measured approximately 40 kN per ground anchor. This results in a displacement of 1.4 mm at position D (close to support A) and 3.3 mm at position C (approx. in the middle of the field). Increasing the load to 77 kN results in a total displacement of 3.0 mm and 7.4 mm at positions D and C, respectively. In the last step, a total load of 95 kN leads to 4.0 mm of deformation close to the support and 9.5 mm of deformation in the middle of the field. When the loading cell is dismantled, the load decreased slightly (approx. 5 kN reduction), accompanied by a recovery of the deformation. Accordingly, the final deformation was 3.7 mm and 8.8 mm.

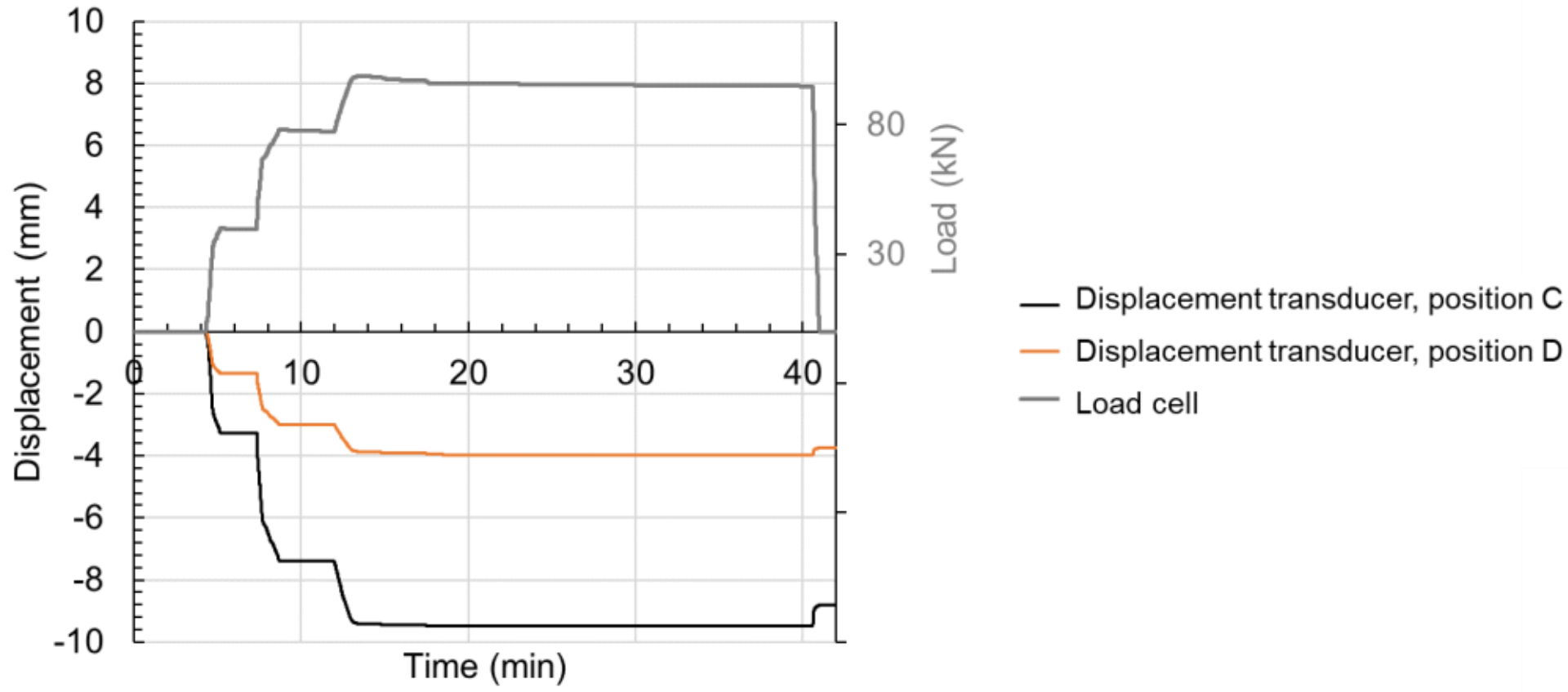

Figure 9. Loading and deformation of the experimental bridge during loading (epoch 3 to 4)

After loading the bridge, where the displacement is investigated for a short period of time (epoch 3 to 4), also the long-term behaviour is examined (epoch 4 to 5). Deformations may occur due to temperature differences and relaxation processes; therefore, single point measurements from different days are not necessarily comparable. For example, over a longer period of time - 20 days in our case - the displacement varies significantly with the environmental temperature, as depicted in Figure 10.

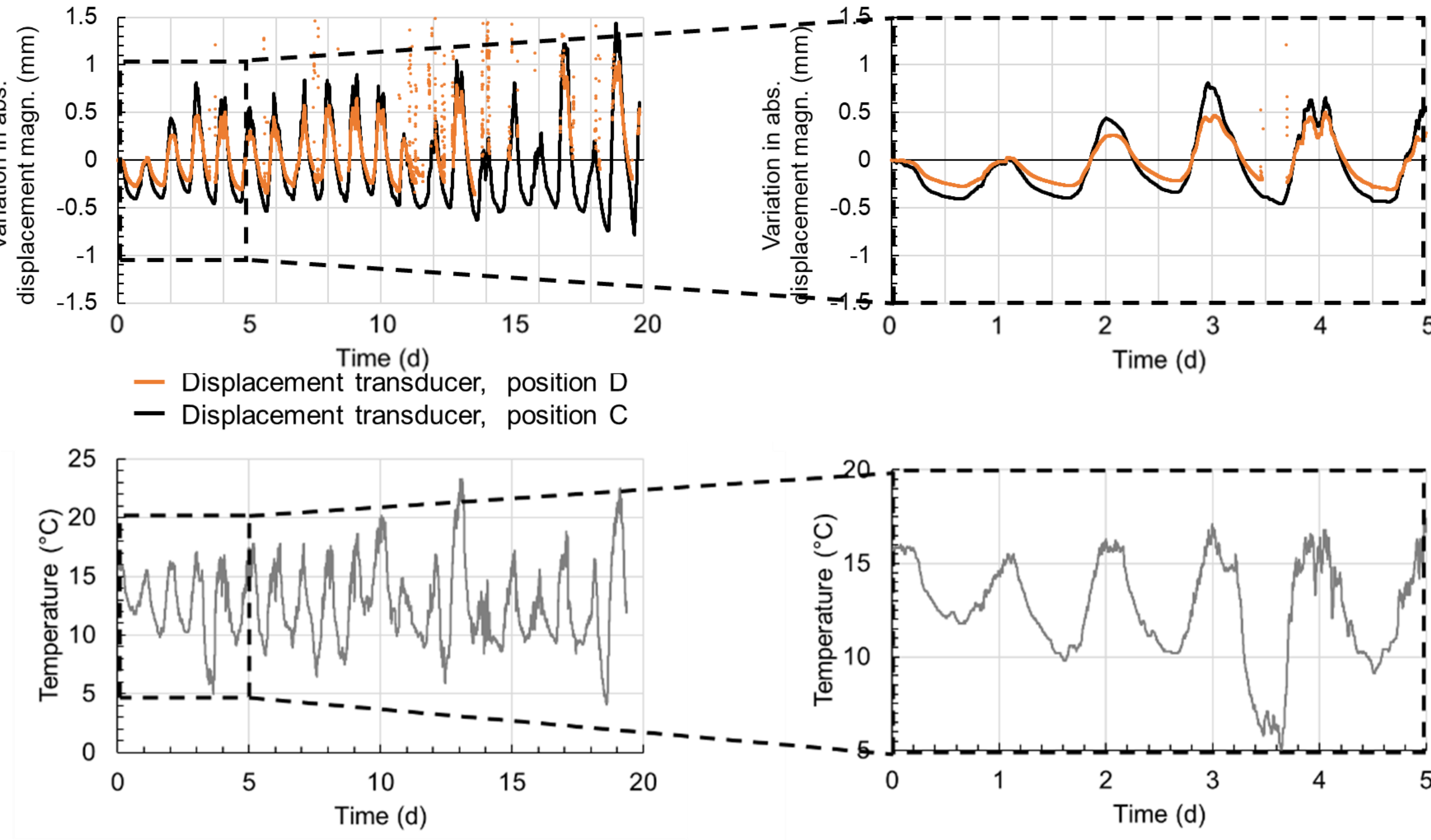

Figure 10. Long-term deformation of the bridge (top) and environmental temperature (bottom) (epoch 4 to 5)

In Figure 10, we show the variation of the absolute displacement magnitude over time to facilitate comparison of the different displacement transducer locations. During the course of the days, the displacement varies according to the change in temperature, with maximum variations of up to 2.3 mm at position C, i.e. approx. in the middle of the field (temperature variation from 4 to 23 °C). This can be mainly attributed to the temperature-related length variation of the steel ground anchors and the corresponding variation of the applied load. The length of the steel anchor decreases as the temperature decreases, resulting in an increased load, and consequently, an increased absolute deformation magnitude. In addition, a slight decrease in the absolute deformation magnitude can be observed over the 20 days, which may be explained by a relaxation of the concrete.

#### 4.1.2 Results of tacheometric deformation measurement

A complete network adjustment with several observations from stations 10, 20, 30, and 40 (cf. Fig. 4) was performed. Afterwards, for each epoch all observations during the experimental period were made from station 10 with orientation to R1 and 40. The measurements were made with the same total station and the atmospheric corrections were entered directly on the instrument, but only one full set of observations was executed. As with the initial network measurement, there were visibility obstructions due to vegetation or wind. It was also shown that the use of different mini prism types is not optimal despite known addition corrections. Despite the influence of wind, slightly different weather conditions and different prisms, a standard deviation of better than 1 mm could be achieved for every epoch.

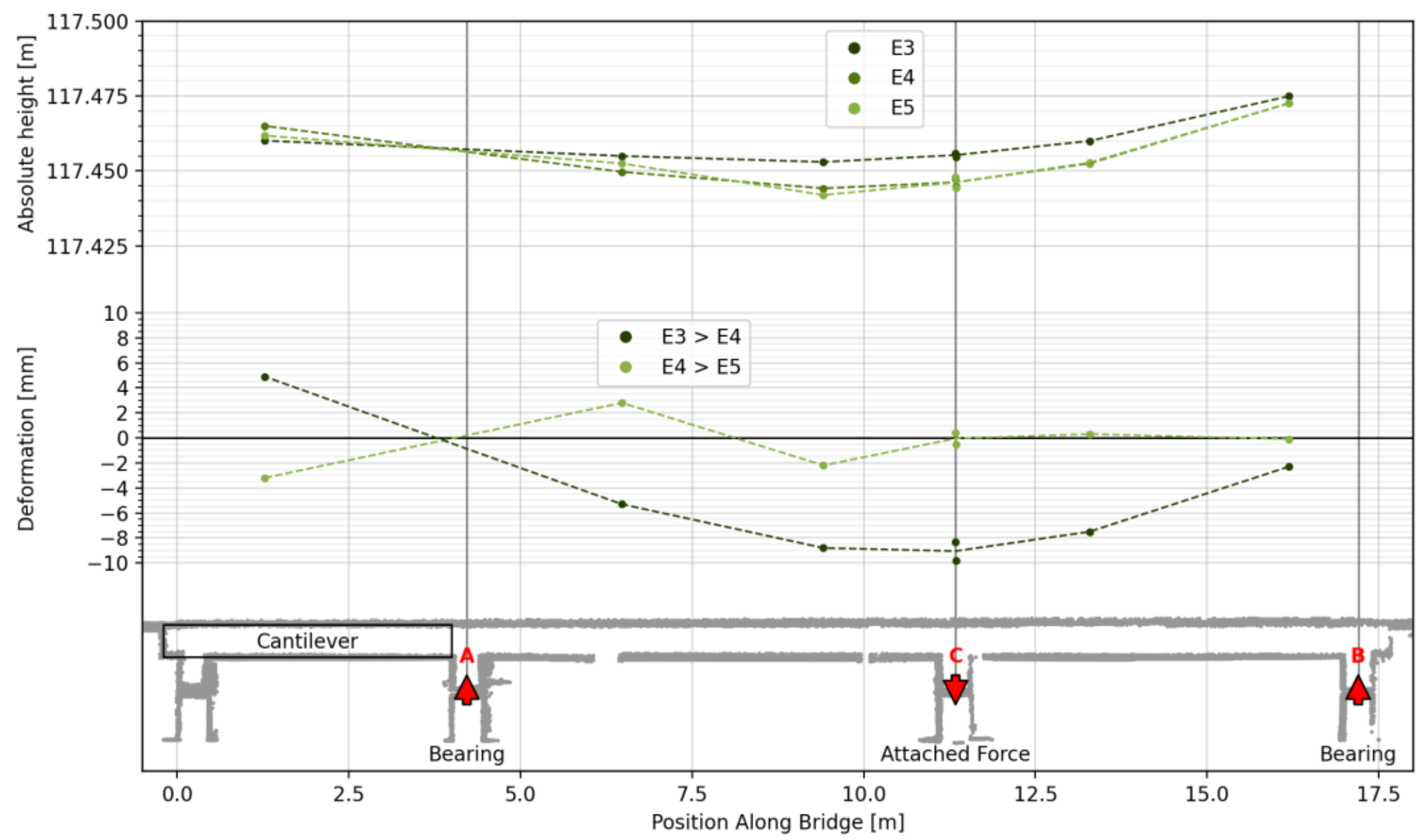

Figure 11. Tacheometry results. Top: absolute heights in epochs 3, 4 and 5. Bottom: Deformation during load and long-term reforming based on differences between epochs.

The elevation values presented in Figure 11 for the bridge's object points are derived from the point-wise observations and epoch-wise calculations described in section 3 and show a rough but very precise point-by-point discretization of the bridge's bending line in the longitudinal axis. An important detail is that at the position of the right end transducers, two tacheometric points were observed, both at the outer edges (North and South) of the deck (c.f. Figure 4: points 22 and 23). It is visible in the plots that those show different motion magnitudes: in the difference of epochs 3 to 4, the northern part got lowered by 8 mm, while in the southern part, it was 10 mm.

#### 4.1.3 Results of TLS measurement

To process the captured profiles, points which did not belong to the underside of the bridge deck needed to be removed from the scan lines. Afterwards, one representative profile line was selected for each epoch to have a “static” scene and to compare the results to the other methods and devices.

The absolute heights of the bottom side of the bridge as measured with the laser scanner can be seen in the upper part of Figure 12. The laser scanner itself has a vertical resolution of 1mm, also there was a high level of noise on the measured values, caused by edge effects and inherent sensor noise. Therefore, all profiles were smoothed, using a Gaussian filter with

σ = 1 cm. The darkest curve shows the height in epoch 3, right before the loading of the bridge. The medium blue line represents the state of epoch 4, directly after the bridge was loaded. Epoch 5, after waiting 20 days for long-term deformation is shown in light blue.

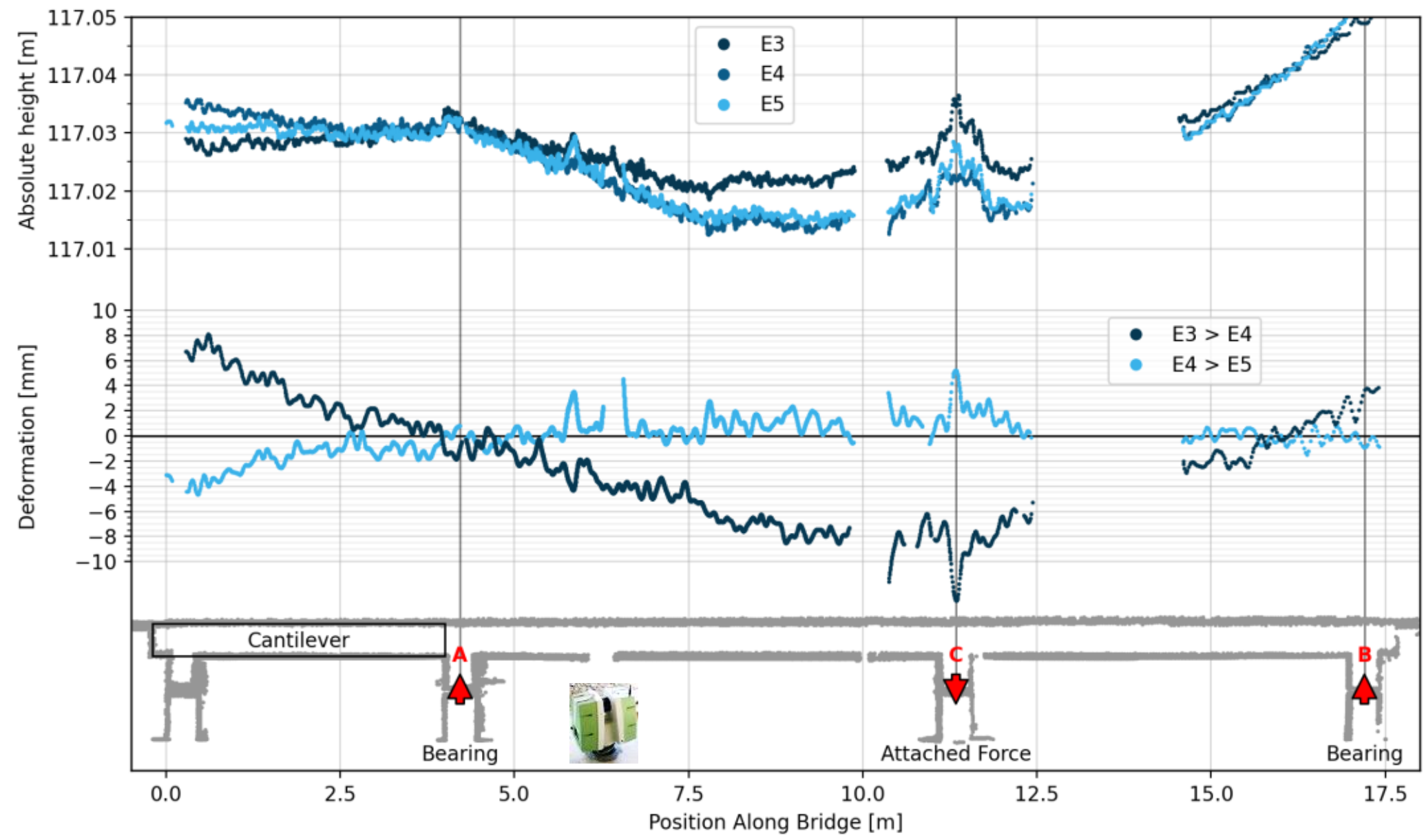

Figure 12**.** Laser scanning results. Top: absolute heights in epochs 3,4 and 5, smoothed by a Gaussian filter with σ = 1 cm. Bottom: Deformation during load and long-term relaxation, smoothed by a Gaussian filter with σ = 3 cm. Deformations were calculated as a difference between two unfiltered height profiles.

At the location of the left support, which is near to the position of the laser scanner, it can be seen that the plots are very well aligned in along the profile and also in vertical direction. The vertical offset between the epochs is approximately zero, as it can be expected at the support. Surface peaks in the profiles are very well in line. The holes on the right side are results of shadowing effects by a crossbeam underneath the deck. Also, it has to be considered that the number of points reduces with increasing horizontal distance from the location of the laser scanner, which is expected considering the measurement principle of this device. A decreasing accuracy with increasing distance is also to be assumed.

To calculate the bending lines between two epochs, the point clouds from the laser scanner had to be referenced to each other. As the horizontal locations of the points differ between the epochs, each epoch was linearly interpolated to a raster width of 1mm. After that, each epoch was further Gaussian filtered with σ = 3 cm to reduce the noise, and finally, the difference was calculated. The smoothed bending line which resulted from the loading process is shown in dark (epoch 3 to 4), and the long-term deformation between epoch 4 and epoch 5 is presented in light blue.

Because of the under-deck structure and measurement noise, the underlying bending line of the bridge during the load cannot be recognized easily. The maximum downward offset is measured at the position of the applied force at position C. The distinct peak at this position results from the concrete structure under the bridge, which is rather uneven at this point of the bridge. At the left support A, there is no offset – as it was assumed – and the cantilever arm has raised. However, there seems to be an error in the measurements near the right support, as the bridge cannot have been lifted there during the load. This effect cannot be explained properly: epochs 4 and 5 are reasonable, also in the right-hand part. This could only mean that there is a problem with the measurement in epoch 3. The maximum deformation at the cantilever arm (difference epoch 3 to epoch 4) is around 6 mm at position 1.5 m, which fits well to the other sensors, so also a systematic height offset in epoch 3 can be excluded.

The light blue line, representing the change in deformation over the long term, i.e. the relaxation between epochs 4 and 5, seems to reflect the motion more realistically. There are only small mean movements smaller than 2 mm in the centre part of the bridge. Only the cantilever arm has gone down slightly by up to 5 mm. A detailed comparison of these bending line results with the other methods of measurement will be found in section 4.2.

#### 4.1.4 UAV-based photogrammetry results

As the photogrammetric reconstruction is carried out in multiple steps, namely the bundle adjustment, the densification and the derivation of a digital elevation model, we will elaborate on those parts in this sequence. All camera parameters got calibrated within the bundle adjustment. Also, the internals got estimated in that step (self-calibration), because we did not have access to a proper lab calibration for the focal length we chose for this project. The lens distortion was modelled according to the Brown model. Since the GCPs were distributed in different heights we were able to reduce the correlation between focal length and exterior orientation where we just used the GCPs. The deviation of GNSS-antenna and the camera centre was not calibrated and RTK information of the drone were used just for flight control. The interested reader can refer to [68] for more details about photogrammetry processes.

Table 2 shows the difference at the on-bridge CPs between the tacheometry measurements and the photogrammetry reconstruction in epochs 3, 4 and 5. Those 3D-points serve as CPs for the respective epoch since between the tacheometric measurement and the UAV-flight no motion is assumed. One can easily see that the horizontal errors are exceeding the desired accuracy of 1 mm, significantly. The highest reconstruction error is almost 8 mm. The reason for these relatively large deviations is of a physical nature regarding the signalisation of points. For the image-based measurements, we directly measured the head of the bolt which was drilled into the concrete; but we used a reflector, which was mounted on the bold for the tacheometric measurements (cf. Figure 1). Although we apply the known height offset, a remaining error comes from the fact that the bolt could not be drilled strictly vertically into the concrete. This fact leads to a de-centric offset which is higher in the horizontal plane than in the vertical. This observation is supported by the observation that for an individual CP and per component, the errors are in the same range, considering the expected errors (standard deviation of tacheometry and photogrammetry around 1 mm). Anyhow, since those points are only used for internal quality checks and not within the bundle adjustment, we decided to go on with the computation of the densified point cloud.

Table 2. On-bridge checkpoint reconstruction errors compared to tacheometric measurements

| CP | Easting Error [mm] | | | Northing Error [mm] | | | Altitude Error [mm] | | |
|---|---|---|---|---|---|---|---|---|---|
| | Epoch 3 | Epoch 4 | Epoch 5 | Epoch 3 | Epoch 4 | Epoch 5 | Epoch 3 | Epoch 4 | Epoch 5 |
| 20 | -1.6 | +0.7 | - 1.4 | +7.0 | +5.2 | +6.7 | +1.3 | - 0.5 | - 0.5 |
| 21 | -2.1 | - 0.3 | - 0.3 | +6.8 | +3.2 | +6.2 | - 2.5 | +0.1 | - 2.7 |
| 22 | +4.5 | +2.2 | +4.5 | +7.7 | +4.4 | +4.8 | - 2.3 | +0.6 | +0.1 |
| 23 | +0.9 | +0.8 | +1.4 | +7.9 | +3.1 | +6.5 | - 1.7 | - 0.4 | - 1.4 |
| 24 | -1.7 | - 0.9 | - 1.1 | +6.7 | +6.3 | +7.0 | - 0.2 | - 0.1 | - 0.2 |
| 25 | +2.4 | +1.1 | +5.4 | +2.4 | +2.3 | +2.2 | - 3.1 | - 1.1 | - 2.2 |
| 30 | +2.6 | - 0.4 | +4.6 | +2.1 | +2.2 | +3.2 | +1.9 | +1.2 | +3.3 |

Figure 13 shows the extracted profiles from all epochs in the top and one can easily see that the horizontal reference between the epochs is very precise. The peaks originate from notches in the surface of the bridge and also the tripods and cables added for our measurements. Epoch 3 is shown in dark red, epoch 4 – directly after applying the load – is seen in medium and epoch 5 in light red.

The bottom part of the figure shows the variation in deformations between the epochs, smoothed by a Gaussian filter with $\sigma = 1$ cm to reduce noise stemming from the above-mentioned artefacts. In the deformation during the load, a perfect reference between the epochs can be recognised as both supports show no vertical deformation. The shape of the bending line is very smooth: One can even notice the point of inflection near the left support A, where the bending line switches from concave to convex.

There are obviously some reference offsets between epochs 4 and 5, as both supports show a value of deformation that is not zero. The general shape is nevertheless plausible up to this point. The next section provides a more detailed comparison of the different types of measurement techniques.

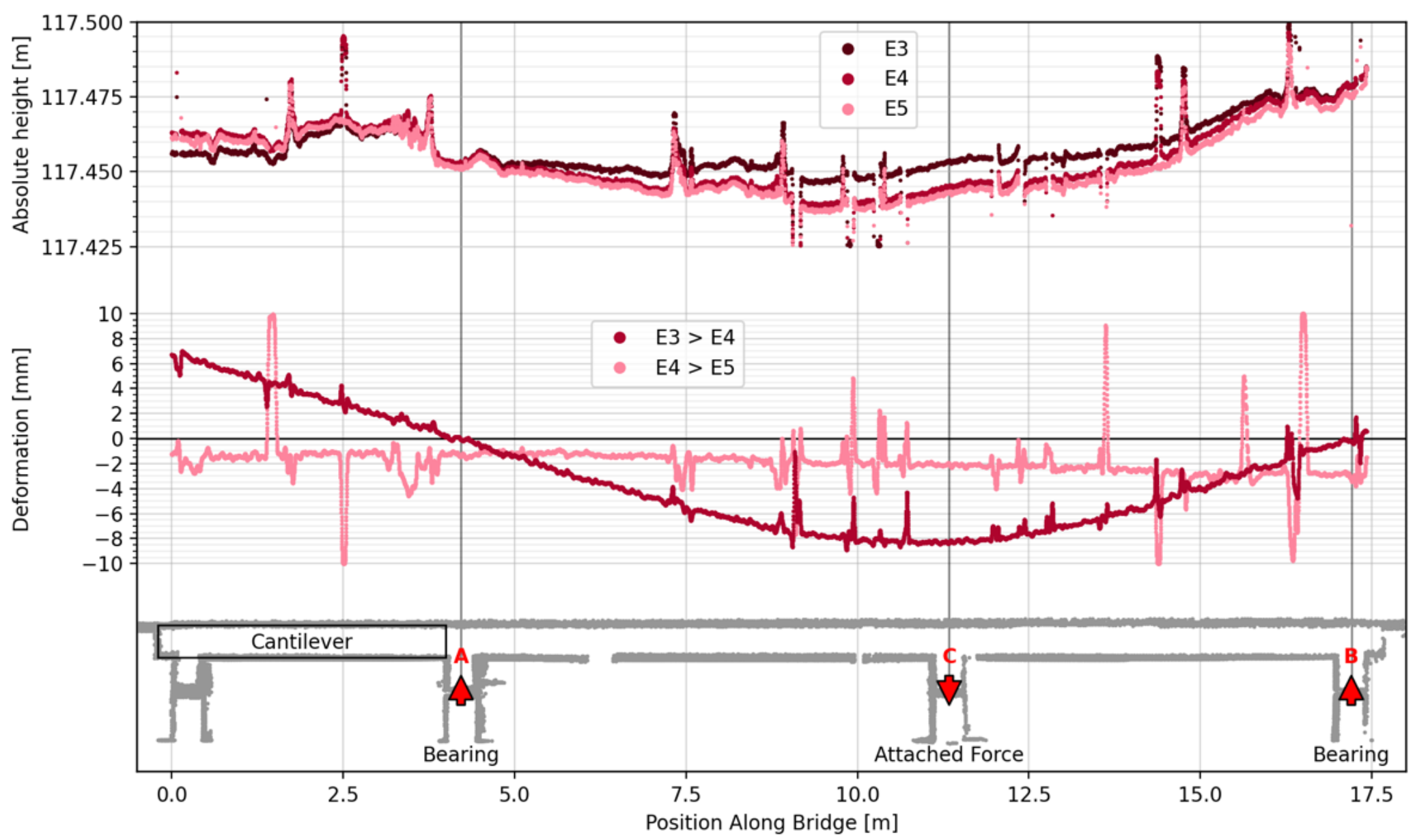

Figure 13. Photogrammetry results. Top: absolute heights in epochs 3,4 and 5. Bottom: Deformation during load and long-term reforming. Bending lines are smoothed by a Gaussian filter with σ = 1 cm.

In this case study, photogrammetric reconstruction serves as the sole method capable of providing not only the deformation along a profile but also as an areal representation. This allows new insights into the deformation, i.e., about bending along cross sections and twisting. In Figure 14, the deformation observed during load application (epoch 3 to 4) is displayed using color-coded projection on an orthomosaic of the bridge. Areas that descended due to the load are shown in red, while green indicates parts of the bridge that got lifted. The support areas are shown in black, indicating no deformation.

While the colour-coding provides a general overview, detailed profiles along any axis can easily be extracted after 3D reconstruction. This provides much more flexibility in the setup, as the axes of interest don’t have to be known in advance.

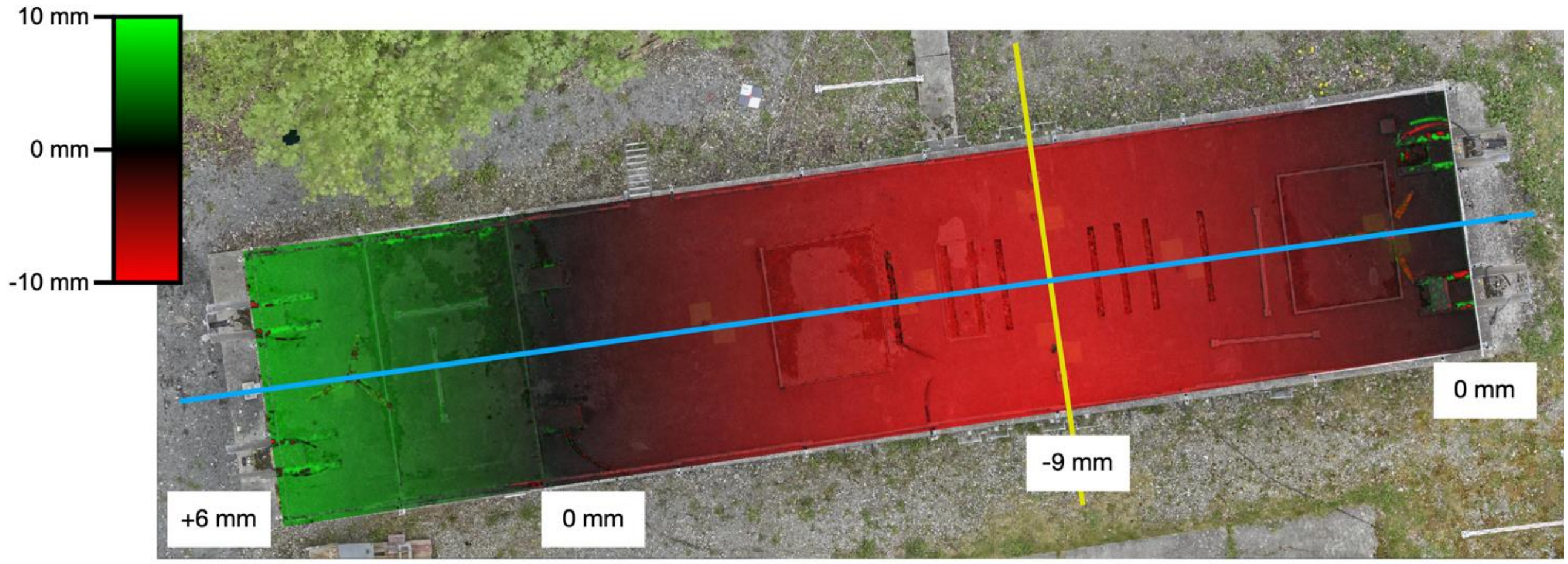

Figure 14. Orthomosaic of “Concerto” overlaid with a plot of the deformation under load as seen from the top according to photogrammetric measurements. Sagged areas are tinted in red, elevated areas in green, no elevation change in black. The bending line is marked in blue and the a cross section at the position of the displacement transducers is marked in yellow .

To demonstrate this ability and to further analyse the photogrammetric result and the behaviour of the bridge, a cross-profile (yellow line in figure 14) through the deformation point cloud (epoch 3 – epoch 4) was extracted orthogonally to the main direction in the position of the transducers. The result is shown in Figure 15.

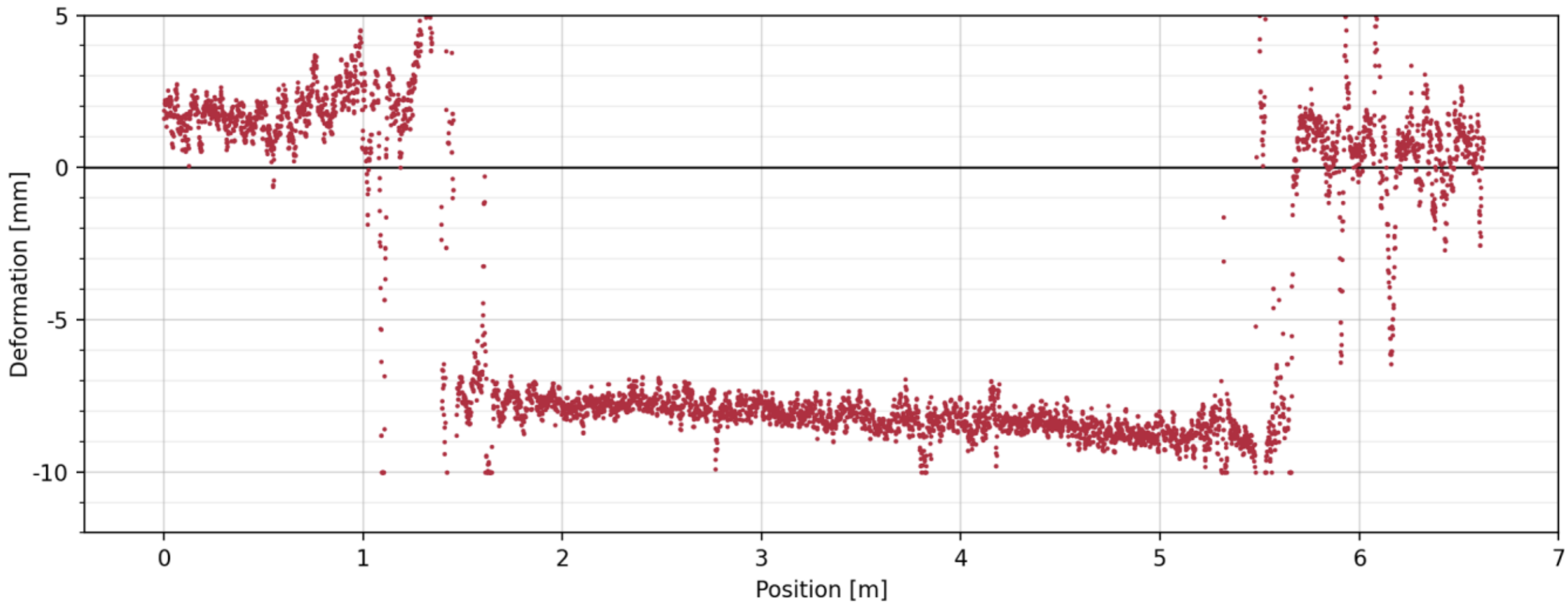

Figure 15. Cross profile from North (left) to South (right) through the epoch 3 to 4 difference point cloud, captured at transducer position C (yellow line in Figure 14).

We can make several interesting observations: the bridge deck was lowered in average by 8 mm, but not constantly: the downward deformation is around 7 mm in the North and 9 mm in South direction. In addition, the outer area seemed to have lifted by 1-2 mm, with a small inclination towards North. This plot is derived from point-to-point distances from the photogrammetric point clouds of the respective epoch. These distances are quite sensitive to small changes in one of the epochs, i.e. if there is vegetation cover. So, in this figure we see some noise (on top of the deck approx. ±1 mm, outside ±2 mm. Especially on the outside we have some outliers in the difference stemming from changes on the surface (vegetation).

## 4.2 Comparison Between different Techniques

Figure 16 shows the change in deformation resulting from load application (between epochs 3 and 4) and the corresponding bending lines as measured with the different methods described in previous sections.

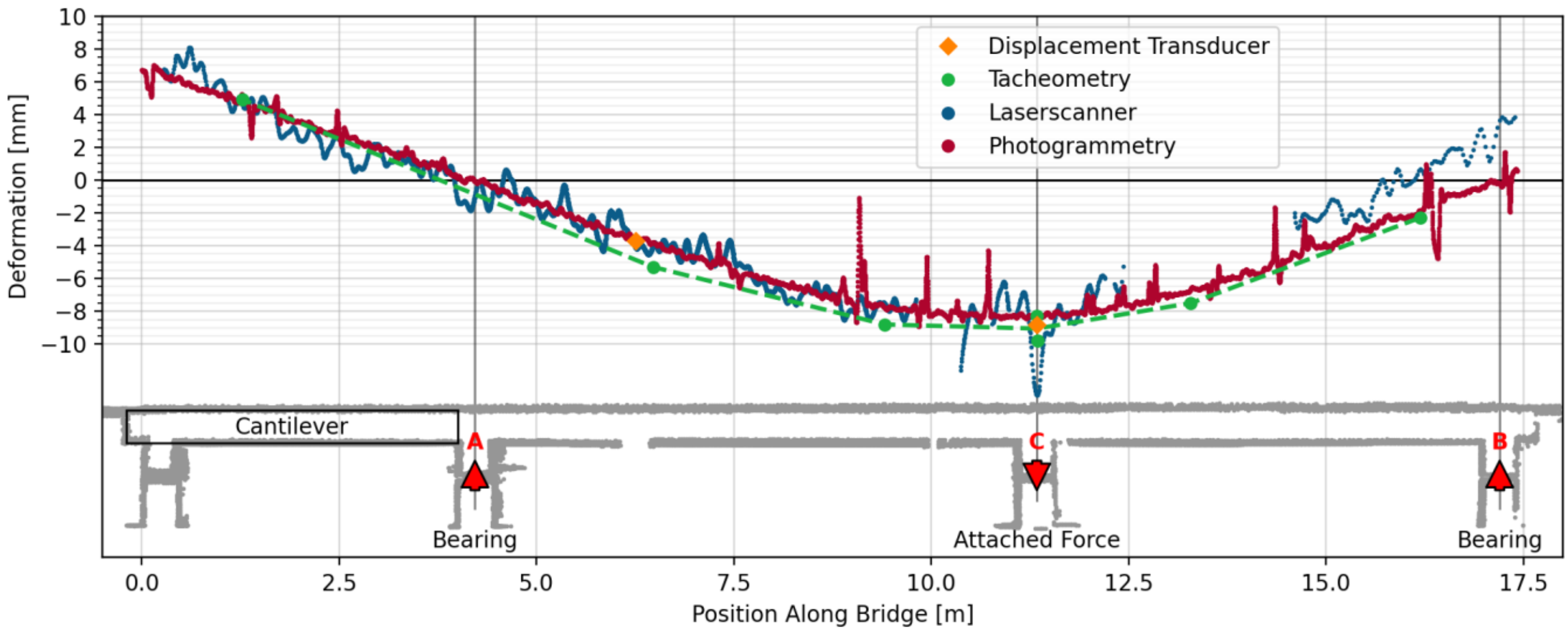

Figure 16. Measured deformation along the centre profile of the bridge between Epoch 3 and 4 (loading).

Due to different capabilities of various measurement methods some of the results cannot be compared directly. Yet **Table** 3 shows that all results of area-based methods can be transferred back to profiles and these again to points, which at least allows an accuracy comparison of all methods at distinct points.

**Table 3.** Evaluation capabilities of the different methods
(green: fully possible per se, red: not possible, orange: see comment)

| Method | Point-based Evaluation | Profile-based Evaluation | Area-based Evaluation |
|---|---|---|---|
| Displacement Transducers | | | |
| Tacheometry | | | |
| Terrestrial Laser Scanning | | | * |
| UAV-based Photogrammetry | | | |

* a 3D area-based scan is possible. However, due to the special geometry of the bridges, in reality it is not optimal to block the entire bridge for laser-scanning with reliable geometry, therefore it was not conducted in this case study.

The two yellow diamonds in Figure 16, representing the displacement transducer results, are assumed to be the most exact measurements and, therefore, considered as ground truth. One can see that the UAV photogrammetry generates good deformation results at these two points, with the difference below 1 mm. The laser scanner observations are disturbed by surface irregularities on the underside of the bridge and noise. However, the overall shape of the graph indicates that the error is in the same range, apart from the unknown effect around support B in epoch 3.

When comparing the ground truth to the tacheometry results, an error of around 1.5 mm is noticeable close to the transducer point at position D. At position C, the error is much lower. At this profile location, there are 2 tacheometry points, one on the northern and one on the southern side of the bridge and the deformation of the centreline is considered to be in the mean of these two values. The uneven behaviour of the two tacheometry points, which are in the outer edges of the deck in the position of the transducer C, was explained above: it indicates a lateral tilt of the deck during the loading phase.

As far as the other total station observations are concerned, it becomes obvious that the bending line resulting from the point connection (green line) shows an offset to the bending lines computed from laser scanning and photogrammetry. Although this shift is around 1 mm, it could be explainable considering the atmospheric condition during the experiment. At each epoch, for tacheometric measurement of the points on top of bridge (except for the two outer ones) reflectors were installed directly above the concrete (cf. Figure 1) and were measured from one station. Due to the hot weather when the experiment was performed, and the severe observation angle, refraction could affect the measurements, systematically but differently for each epoch. This can be verified when there is no such a systematic shift on the point where reflectors are installed on the tripod (cf. Figure 16 and Figure 1). Moreover, uncertain signalisation of the bolts, fixed in the concrete might affect the observations.

On the right support, however, only the UAV-photogrammetry results fit the assumption that there should not be any vertical deformation at this point. The laser scanner result shows odd behaviour in the area around this support. While the absence of results between 12.5 m and 14.5 m is clearly caused by the shadowing by a load-bearing structure, the offset in deformation of around 3 mm in the later areas cannot be explained. The measurement in epoch 3 causes it, but one also has to consider that the error of up to 3 mm at a distance of around 10 m is exactly within the accuracy range the manufacturer specified.

Compared to tacheometry, laser scanning and UAV-photogrammetry generate results with a much higher spatial resolution. The deformation can only be computed indirectly through differencing height observations from two epochs. Such differences are affected by noise resulting from individual changes on the surface which are not caused by deformation, and by uncertainties in assigning corresponding points for differencing. Therefore, such computations result in noisy bending lines, which must be smoothed with a suitable low-pass filter to be comparable at a distinct location.

Figure 16 shows a much smoother graph for the UAV photogrammetry, although this was only filtered by a Gaussian filter with $\sigma = 1$ cm. In contrast, the bending line resulting from laser scanner measurements was smoothed with $\sigma = 3$ cm and is still disturbed by more peaks. It has to be considered that both methods observe different sides of the bridge, as explained.

## 5 Conclusions

The objective of the work carried out in this research was to compare and analyse different sensing techniques for in-situ SHM tasks. To this end, we equipped a research bridge with several sensors and applied well-defined loads in different epochs. Though our setup had some limitations concerning point signalisation for tacheometry measurements we gained valuable insights.

The UAV-based photogrammetric block was adjusted using tachymetric network measurement in stable areas around the bridge. Despite some remaining uncertainties, the general bending behaviour resulting from the applied force matches the transducer measurements within a difference of better than 1 mm. This confirms the assumption that the bridge does not deform at the support. On the other hand, the remaining uncertainties within the UAV block in our experiments do not allow conclusions about the long-term stability (epochs 4 to 5). Thus, it was impossible to quantify the effect of temperature-induced deformation. Total station measurements indicated a tilt in cross-direction, possibly caused by uneven force application. This behaviour was verified by the cross-section derived from the UAV-based point clouds. This example confirms our hypothesis that areal measurements derived from the UAV point cloud can give much more comprehensive insights into the bending behaviour and can thus complement classical point measurements. Detailed profiles along any axis can easily be extracted after 3D reconstruction, providing much more flexibility in the setup, as the axes of interest do not have to be known in advance.

The profile obtained by laser scanner measurements underneath the deck confirmed the overall bending behaviour. However, due to interfering surface structures, a large filter had to be applied, which further lowered the accuracy of the result. In addition, large offsets in epoch 3, after load, led to unexplainable effects in the eastern part of the structure. On the other hand, the nominal accuracy of the used laser scanner does not allow further analysis of this effect. It should be highlighted that our experiment was performed on a test bridge. Since the height of this bridge is less than 1.5 m, the TLS placement affected the quality of laser scanning method which could be avoided in real scenarios where there is enough space underneath the bridges.

Though our setup had some limitations concerning point signalisation for total station measurements and laser scanning setup, we demonstrated that using UAV-based photogrammetric image blocks at large infrastructure objects can give valuable insights into the overall deformation behaviour. It also should be considered that dynamic behaviour cannot be monitored with this method. On the other hand, holistic deformation patterns can be derived from UAV-based photogrammetry and might give insight into understanding structural behaviour, for instance, in long-term monitoring tasks.

Due to the displacement of the points between the epochs, more attention is necessary for deformation monitoring tasks. Care has to be taken regarding a proper geodetic network measurement and point signalisation to derive appropriate deformation information. With more sets of observations or more total station positions, we could have more control on the network and have a compromise between time and reliability of the results. Moreover, TLS data capture geometry (incidence angle, mixed pixels, etc) and the object’s geometric complexity should be considered to avoid extra inaccuracies. Moreover,

**Acknowledgments:** The Project “Optical 3D-Bridge-Inspect” is funded by the Deutsche Forschungsgemeinschaft (DFG, German Research Foundation) – Project Number 501682769 as part of DFG’s Priority Programme 2388 “Hundred plus”. The authors would also like to acknowledge support by the Open Access Publication Funds of the Technische Universität Braunschweig.